\documentclass[conference]{IEEEtran}
\IEEEoverridecommandlockouts
\usepackage{cite}
\usepackage{amsmath,amssymb,amsfonts}
\usepackage{{booktabs}}
\usepackage{multicol}
\usepackage{multirow}
\usepackage{array}
\usepackage{hyperref}
\usepackage[ruled,linesnumbered]{algorithm2e}
\usepackage{graphicx}
\usepackage{textcomp}
\usepackage{xcolor}
\usepackage{tikz}

\def\BibTeX{{\rm B\kern-.05em{\sc i\kern-.025em b}\kern-.08em
    T\kern-.1667em\lower.7ex\hbox{E}\kern-.125emX}}

\usepackage{xspace}
\usepackage{soul}
\usepackage{acronym}

\newcommand{\etal}{\textit{et al.}\xspace}

\newcommand{\ie}{\textit{i.e.},\xspace}

\acrodef{ml}[ML]{Machine Learning}
\newcommand{\ml}{\ac{ml}\xspace}

\acrodef{fl}[FL]{Federated Learning}
\newcommand{\fl}{\ac{fl}\xspace}

\acrodef{fedavg}[FedAvg]{Federated Averaging}
\newcommand{\fedavg}{\ac{fedavg}\xspace}

\acrodef{fedopt}[FedOpt]{Federated Optimization}
\newcommand{\fedopt}{\ac{fedopt}\xspace}

\acrodef{cfl}[CFL]{Clustered Federated Learning}
\newcommand{\cfl}{\ac{cfl}\xspace}

\acrodef{ocfl}[OCFL]{One-Shot Clustered Federated Learning}
\newcommand{\ocfl}{\ac{ocfl}\xspace}

\acrodef{iot}[IoT]{Internet-of-Things}
\newcommand{\iot}{\ac{iot}\xspace}

\acrodef{dp}[DP]{Differentional Privacy}

\acrodef{he}[HE]{Homomorphic Encryption}

\acrodef{smpc}[SMPC]{Secure Multi-Party Computation}

\begin{document}
\newpage
\onecolumn
\noindent\makebox[\linewidth]{}
20XX IEEE Personal use of this material is permitted.  Permission from IEEE must be obtained for all other uses, in any current or future media, including reprinting/republishing this material for advertising or promotional purposes, creating new collective works, for resale or redistribution to servers or lists, or reuse of any copyrighted component of this work in other works.
  DOI: \href{https://doi.org/10.1109/BigData62323.2024.10825763}{10.1109/BigData62323.2024.10825763}.
\newpage
\twocolumn

\title{One-Shot Clustering for Federated Learning\\
\thanks{ \textbf{Acknowledgements:} This work was supported by European Union in projects \emph{LeADS} (GA no. 956562) and \emph{CREXDATA} (GA no. 101092749), NextGenerationEU programme under the funding schemes PNRR-PE-AI scheme (M4C2, investment 1.3, line on AI) FAIR (Future Artificial Intelligence Research), NextGenerationEU – National Recovery and Resilience Plan (Piano Nazionale di Ripresa e Resilienza, PNRR) – Project: “SoBigData.it – Strengthening the Italian RI for Social Mining and Big Data Analytics” – Prot. IR0000013 – Avviso n. 3264 del 28/12/2021”. This work was also funded by the European Union under Grant Agreement no. 101120763 - TANGO. Views and opinions expressed are however those of the author(s) only and do not necessarily reflect those of the European Union or the European Health and Digital Executive Agency (HaDEA). Neither the European Union nor the granting authority can be held responsible for them.
}}

\author{\IEEEauthorblockN{1\textsuperscript{st} Maciej Krzysztof Zuziak}
\IEEEauthorblockA{\textit{KDD Lab} \\
\textit{National Research Council of Italy}\\
Pisa, Italy \\
https://orcid.org/0000-0003-4297-4973}
\and
\IEEEauthorblockN{2\textsuperscript{nd} Roberto Pellungrini}
\IEEEauthorblockA{\textit{KDD Lab} \\
\textit{Scuola Normale Superiore}\\
Pisa, Italy \\
https://orcid.org/0000-0003-3268-9271}
\and
\IEEEauthorblockN{3\textsuperscript{rd} Salvatore Rinzivillo}
\IEEEauthorblockA{\textit{KDD Lab} \\
\textit{National Research Council of Italy}\\
Pisa, Italy \\
https://orcid.org/0000-0003-4404-4147}}

\maketitle

\begin{abstract}
\fl is a widespread and well-adopted paradigm of decentralized learning that allows training one model from multiple sources without the need to directly transfer data between participating clients. Since its inception in 2015, it has been divided into numerous sub-fields that deal with application-specific issues, be it data heterogeneity or resource allocation. One such sub-field, \cfl, is dealing with the problem of clustering the population of clients into separate cohorts to deliver personalized models. Although few remarkable works have been published in this domain, the problem is still largely unexplored, as its basic assumption and settings are slightly different from standard \fl. In this work, we present \ocfl,  a clustering-agnostic algorithm that can automatically detect the earliest suitable moment for clustering. 
Our algorithm is based on the computation of cosine similarity between gradients of the clients and a temperature measure that detects when the federated model starts to converge.
We empirically evaluate our methodology by testing various one-shot clustering algorithms for over thirty different tasks on three benchmark datasets. Our experiments showcase the good performance of our approach when used to perform \cfl in an automated manner without the need to adjust hyperparameters.
\end{abstract}


\section{Introduction}
The recent decade has witnessed exponential growth in the number of available data sources, mostly due to the expansion of \iot products and the popularisation of large data storing facilities that allow for the storage of unprecedented amounts of information. To leverage such an infrastructure, new paradigms of learning emerge, allowing data engineers and scientists to train models distributively without the need to transfer the data into a central place - which often would be unrealizable due to the sheer volume of stored information.

One such modern approach to decentralized \ml is \fl, a paradigm first introduced by McMahan \cite{b22} where the local models are trained locally, and the final (global) model is aggregated from the parameters of the local equivalents. Since the baseline \fl algorithm has been introduced under the name of \fedavg, the field has undergone recent developments with algorithms such as Adaptive Federated Optimization (FedOpt) \cite{b23}, SCAFFOLD \cite{b24}, and FedProx \cite{b40} addressing the specifying challenges that may arise in a federated environment.

The task of clustering clients into several conglomerations to train separate (personalized) models is one of such specific challenges that may arise, especially in the cross-silo setting, where the number of participants is limited to dozens rather than thousands and the institutional nature of the participants may be required to deliver personalized solutions, at least at the granularity of the regional level. One example of such a situation may be training a global risk model for the insurance companies, where the major differences in the local market are so vast that the final (global) model is inoperable to any of the parties interested. In the majority of such situations, companies may not know a) that there exists some form of incongruence and b) what grouping would be required to deliver models of satisfactory quality. The sub-field of \cfl addresses such issues, applying clustering on the population of clients to deliver several personalized models that still retain some level of generalizability.

In this work, we approach the issue of fast and reliable \fl clustering that is automatically performed in the first few rounds of the training. Since many works in the field of \cfl employ clustering algorithms on the matrices of pairwise cosine distance (or cosine similarity), our main interest was focused on the generalization of the presented procedure with an improvement of the clustering efficiency, especially clustering correctness. We, therefore, investigate whether it is possible to automatically detect the suitable clustering moment and perform efficient one-shot clustering at an early stage of the process without specifying any hyperparameters. Moreover, we turned our attention to the scenario of medium heterogeneity, where the clients differ only slightly in data-generating distributions.

We propose a clustering-agnostic algorithm that is free of any hyperparameters and clusters at the first sign of the (global) model's convergence. Our approach, called \ocfl, can be used to perform a reliable clustering in the first rounds of the training without the need to adjust either the number of clusters or precise clustering criterion. As evidenced by the experimental section, when combined with density-based clustering (e.g. HDBSCAN \cite{b19} or Mean-Shift \cite{b20}) it achieves better results than State-of-the-Art \cfl algorithms when it comes to clustering performance and personalization, while retaining comparable levels of generalizability.

Since our proposition is clustering algorithm agnostic, it can be combined with most of the existing clustering algorithms. Because of that, we share our experimental code base that can serve as a foundation for further experimentation.\footnote{\href{https://github.com/MKZuziak/Clustered-Federated-Learning}{Link to the GitHub repository}} All the experiments and their visual representation can be re-created using code deposited at the GitHub Repository shared with this article. \footnote{\href{https://github.com/MKZuziak/OCFL_IEEE_BigData2024}{Link to the GitHub repository}}.
We also provide a formalization of the data distribution process in the evaluation of our method. While non-i.i.d. data distributions are an obstacle for many \fl algorithms, they are easier to cluster than distributions with more subtle but still noticeable differences. In this article, we experiment with distributions of medium heterogeneity to evaluate the real effectiveness of the similarity-based \cfl algorithms. 
Our paper is organized as follows: in Section \ref{sec:related} we indicate the literature most relevant to our work, in Section  \ref{sec:methodology} we describe in detail our methodology, in Section \ref{sec:experiments} we present our experimental evaluation with a detailed discussion on the results, in Section \ref{sec:considerations} we add some additional comments on the general impact of our work and finally in Section \ref{sec:conclusions} we conclude our work.

\section{Related Work}
\label{sec:related}
\subsection{Federated Learning} \label{RelatedWork: FL}
\fl was first introduced in the work of \cite{b22} as a decentralized optimization method, where the training is distributed across several clients where each client shares only the parameters of the model trained on the local model, not disclosing its local (personal) dataset directly. The basic method of solving this problem, \fedavg (presented in \cite{b22}) assumes averaging values of the local parameters to arrive at the general global model. In the following years, different additional solutions to that problem were presented. \fedopt \cite{b23} generalizes the \fedavg algorithm, allowing one to introduce adaptive optimization into the process. Among others, SCAFFOLD \cite{b24} and FedProx \cite{b40} tackle the issue of data heterogeneity, limiting the gradient variability and thus stabilizing the convergence process. As \fl became an important area of research, several surveys have tried to summarize the advancements made in the last decade of research \cite{b25, b26}. In this paper, we assume a horizontal learning scenario. In contrast to the vertical setting \cite{b27}, the horizontal one assumes that the sample space is disjoint, while the feature space is shared among all the participants. Implementing our clustering algorithms, we employ a \fedopt baseline \cite{b23}. Given a set of disjoint datasets $D = \bigcup_{i \in P}D_i$, where each dataset $D_i = \{\boldsymbol{x}_i, y_i\}_{i=1}^k$ is generated according to a certain probability distribution, i.e. $D_i \sim \varphi_i$, the \fl task is to optimize global hypothesis function using aggregation of the locally trained functions, i.e. $ argmin_{h \in \mathcal{H}} R_{ERF}(h, D) \approx \frac{1}{|P|} argmin_{h \in \mathcal{H}} R_{ERF}(h, D_i)$, where $P$ is the size of the population. In order to do that, we employ the \fedopt algorithm presented in \cite{b23}, which is a well-accepted baseline.

\subsection{Clustering in FL} \label{RelatedWork: FCL}
\cfl is currently divided into two distinct approaches, which we call here \textit{hypothesis-based clustering} and \textit{parameters-based clustering}. The primer is based on the similarity of the hypothesis function \cite{b3, b9} and relies on matching clients with the most similar objective functions. The former relies on the similarity of the local models' parameters, as they can disclose information about the local data generating process \cite{b1, b7, b2, b4, b10, b8}. In this paper, we take a closer look at the second group of methods as we propose a natural extension of methods based on the similarity between the local sets of parameters. 

Much of the work in the \textit{parameters-based clustering} domain relies on comparing weights or gradients provided by the local models. Two main aspects that distinguish particular approaches are the time of the clustering and the clustering algorithm. Briggs \etal \cite{b1} propose using hierarchical clustering based on models' parameters. Their work implies fixing a clustering round \textit{a priori} together with a choice of a proper linkage criterion and its corresponding hyperparameters (mainly, distance threshold). The work of Sattler \etal \cite{b7} relies on a cosine similarity-based bipartioning, where the clustering is based on the cosine similarity between different sets of parameters. Since the clustering is postponed until the (global) FL model converges to a stationary point, it is deemed a postprocessing method. The work of \cite{b2} employs an Euclidean distance of Decomposed Cosine Similarity (EDC) to cluster clients. Authors of \cite{b5} mix distance comparison with client-side adaptation (varying number of local epochs) and weighted voting. In the \cite{b10}, authors use pairwise similarity to measure the statistical heterogeneity of the local datasets. In \cite{b8}, authors experiment with the use of cosine distance and client-side code execution to correct possible erroneous clustering results (by individual client drop-out).

Our work resembles propositions made by Sattler \etal \cite{b7} and Briggs \etal \cite{b1}. Hence, those two articles serve as our natural comparison point. By experimenting with different types of splits (with a heavier emphasis on the \textit{feature skew} and \textit{label skew}), we hope to further expand their methods in a more challenging (clustering-wise) environment and automatically identify the earliest possible moment for performing one-shot clustering. Because our method falls within the \textit{parameters-based clustering}, it also can serve as a possible augmentation for methods presented in \cite{b2, b5, b8} and \cite{b10}.

\section{Methodology}
\label{sec:methodology}
\subsection{Data Generating Processes} \label{methodology: data_process}
Simulating a divergence in data-generating processes that resembles the naturally occurring differences in data distributions is a challenge that is approached in various ways. Authors of \cite{b8, b5} adopt label distribution and dataset distribution skew, with \cite{b8} additionally experimenting with differences in feature distribution. Both \cite{b7} and \cite{b1} employ label swaps to stimulate incongruence in data distribution. In the \cite{b6}, authors follow the methodology presented in \cite{b11} using Dirichlet distribution to simulate a heterogeneous environment. Additionally, FEMNIST \cite{b1, b10} and Fed-Goodreads \cite{b10} are employed in some papers.

Considering that there is no consensus on how to simulate naturally occurring clusters of users in FL, we have adopted a precise methodology focused on the data-generating processes that could potentially arise in a natural environment. This formalization allows one to develop various complicated scenarios while retaining a firm methodology of generating the datasplits. Throughout this article, we use $D \sim \varphi$ to denote the dataset that is sampled from a certain data-generating process $\varphi$. 

Using this notation, we start from the notion of multiple data generating distributions $\Phi = \{ \varphi_1, \varphi_2, \cdots, \varphi_n \}$, where each local dataset is either sampled from the probability distribution $\mathcal{D}_k \sim \varphi_i(\boldsymbol{x})$ or is based on the conditional probability distribution $\mathcal{D}_k \sim \varphi_i(y|\boldsymbol{x})$. The goal of the \cfl task is to assign clients to clusters according to their original data generation mechanism. In our setting, this is reflected on the cluster level, where the population of clusters $C = \{c_1, c_2, \cdots, c_n \}$ is attributed to individual data generating distributions by a bijective function $\mathcal{I}: C \longrightarrow \Phi$, i.e. each cluster corresponds to one data-generating distribution.\footnote{This can be generalized beyond the bijective relationship. However, without the loss of generality, we assume here that $\mathcal{I}$ is a bijective function.} The client-cluster correspondence is established by surjective function $\mathcal{T}: C \longrightarrow \hat{P}$, where $\hat{P}$ is a subset of the whole set of clients, i.e. $\hat{P} \subseteq P$.

The type of interactions and overlaps between clusters are fundamental to clustering algorithms. Since $\varphi_i: \mathcal{A} \longrightarrow \mathbb{R}$ is a probability distribution, the difference will be connected to how we define the $\sigma$-algebra input space $\mathcal{A}$. In the case of this article, we will deal mostly with discrete and finite cases, hence $\mathcal{A} = \{ a_1, a_2, \cdots, a_n \}$ where $a_i$ is the $i^{th}$ class belonging to a certain distribution. Given two data-generating distributions, $\varphi_i$ and $\varphi_j$ with respective set-theoretic supports, we indicate with $A_i \cap A_j = \emptyset$ a non-overlapping case, i.e., no common data points. On the other hand, we indicate the overlapping case with $\mathcal{A}_i \cap \mathcal{A}_j = \mathcal{A}_{ij}, \quad$ such that $\mathcal{A}_{ij} \neq \emptyset$. 

When compared with the taxonomy of non-identical client distributions presented in \cite{b25}, our methodology allows for a rigid expression of a \textit{feature distribution skew} (where $\varphi_i(\boldsymbol{x})$ vary across clients, with $\varphi_i(y|\boldsymbol{x}) = \varphi_j(y|\boldsymbol{x}) \quad \forall i, j \in P$) and \textit{label distribution skew } (where $\varphi_i(y)$ vary across clients, with $\varphi_i(\boldsymbol{x}|y) = \varphi_j(\boldsymbol{x}|y) \quad \forall i, j \in P$), additionally combined with \textit{quality skew} and the non-independence of the feature and label space (shared examples). We do not consider \textit{concept drift} and \textit{concept shift} in this paper. While important for studying the effect of convergence in highly heterogeneous environments, feature and label distributions are the most difficult to capture because the model may differ subtly while still benefiting from clustering. On the other hand, it is implied that the case of extreme heterogeneity impacts learning to the utmost level, hence being easier to detect - a fact proven by previous authors published on the topic \cite{b1, b3, b7}. In the case of extreme heterogeneity it may be also pointless to cluster nodes, as generalization and personalization performance of the models would change only slightly.

\subsection{Hypothesis Function Similarity} \label{methodology: function_similarity}
Accessing the information about the local distribution is impossible, as it would nullify the basic assumption underlying the \fl. However, some information about the status of the learning can be derived from the model's parameters that reflect a local hypothesis function. For each cluster $C_i \in C$, there exists a hypothesis function $h_{C_i}$ that minimizes the empirical risk function on a dataset sampled from the corresponding data-generating distribution, i.e. $h_{C_i}^* = argmin_{h \in \mathcal{H}} R_{ERF}(h, D_i) $, where $D_i \sim \varphi_i$. Every client belonging to a certain cluster is training their own hypothesis function $h_{p_i}$ that minimises the local risk function. The hypothesis function $h_{C_i}^*$ is not guaranteed to outperform the local hypothesis function $h_{p_i}$ in every case since it depends on the data stored on the local client. However, given the environment, $h_{C_i}^*$ will be the function that minimizes the empirical risk for the whole cluster. One illustrative example may be connected to a small sample size of the local dataset. If the client is in possession of only a few samples, an overfitted local hypothesis function may outperform the best cluster hypothesis function. However, as the client will procure new samples, the best cluster hypothesis function will perform better than the overfitted local function.

The similarities of local loss functions are used by \cite{b3, b9} to cluster the population of clients into a predefined number of clusters. This approach suffers from a high computational cost (as each client must test out each possible hypothesis function to find the one that minimizes the local risk) and the necessity to know \textit{a priori} the structure of the population, \ie the number of total clusters. However, one could note that the hypothesis function $h_i$ is parameterized by the n-dimensional tensor $\boldsymbol{\Theta}$.\footnote{The dimensions of the tensor will depend on the used model. In the case of conventional $n$ layered feedforward neural networks, the weights are often represented in $n-1$ separate matrices of size $a \times b$, where $b$ is the number of neurons in the layer $j$ and $a$ is the corresponding number of neurons in the $j + 1$ layer. This can be represented as a three-dimensional tensor $\Theta \in \mathcal{R}^{n - 1} \times \mathcal{R}^{a} \times \mathcal{R}^{b}$. Without the loss of generality, we assume that $\Theta \in \mathcal{R}^n$ or $\Theta \in \mathcal{R}^n \times \mathcal{R}^m$ where $n, m \in \mathcal{Z}$.} The flattened one-dimensional version of this tensor, i.e. $\boldsymbol{\hat{\Theta}} = Vec(\Theta)$, can be compared with other parameterized hypothesis functions without the need to directly assess the hypothesis function performance on the local dataset - the method used in \cite{b1, b2, b7, b8}. 

The current state of the art does not provide answers to when the clustering could be initialized at the earliest. To develop an intuitive understanding of the problem, let $C_s(\boldsymbol{\hat{\Theta}}_i,\boldsymbol{\hat{\Theta}}_j)$ denote \textbf{cosine similarity} between vectors of parameters $\hat{\Theta}_i$ and $\hat{\Theta}_j$ and $C_d(\boldsymbol{\hat{\Theta}_i}, \boldsymbol{\hat{\Theta}_j)} = 1 - C_s(\boldsymbol{\hat{\Theta}_i},\boldsymbol{\hat{\Theta}_j})$ denote a \textbf{cosine distance}. Subsequently, define $\boldsymbol{\Gamma} \in \mathcal{R}^{m \times m}$ symmetric matrix that captures the similarity between parametrization of a hypothesis function provided by all the sampled clients. The $(i,j)$ entry of this matrix will be defined as:

\begin{equation}\label{eq: Gamma_Matrix}
    \gamma_{(i, j)} = \gamma_{(j, i)} = C_d(\boldsymbol{\hat{\Theta}}_i,\boldsymbol{\hat{\Theta}_j})
\end{equation}
Matrix $\boldsymbol{\Gamma}$ captures the learning divergence between the sampled nodes. To obtain a single-valued description of the current convergence progress, one can calculate the \textit{Frobenius Norm} of $\boldsymbol{\Gamma}$, i.e.:

\begin{equation}\label{eq: Gamma_Energy}
    ||\boldsymbol{\Gamma}||_p = ||vec(\boldsymbol{\Gamma})||_p = (\sum_{i=1}^m \sum_{j=1}^n |\gamma_{i,j}|^p)^{\frac{1}{p}}
\end{equation}

Note that if $\boldsymbol{\Gamma}$ is an $n \times n$ square matrix, then:

\begin{equation}\label{eq: Energy_Bounds}
    0 \leq ||\boldsymbol{\Gamma}||_p \leq \sqrt[\uproot{1} p]{n(n-1)2^p}
\end{equation}

Let \textbf{Clustering Temperature} be defined as a function mapping divergence matrix $\Gamma$ to $\mathbb{R}$:

\begin{equation}\label{eq: Temperature}
    T(\boldsymbol{\Gamma}) = \frac{||\boldsymbol{\Gamma}||_p}{\lambda}
\end{equation}

This implies that the definition of temperature introduced here will be based on the $p-norm$ of the \textbf{Similarity Matrix} and the scaling constant $\lambda$. Equation \eqref{eq: Energy_Bounds} provides us with some guidance on how to choose the appropriate scaling constant. If we let $\lambda = \sqrt[\uproot{1} p]{n(n-1)2^p}$ and substitute that back into Equation \eqref{eq: Temperature}, we obtain a bounded indicator of the current state of the divergence in the system.

\subsection{Clustering Procedure}
Our proposed algorithm is based on the empirical observations of the \textbf{Clustering Temperature} Function (Equation \eqref{eq: Temperature}). It is expected that the function $T$ will rise at the beginning of the training - and then steadily drop as the convergence progresses across the rounds. Using this observation (that is presented in more detail in the subsection \ref{experiment: temperature_function}), we present an \textit{\textbf{\ocfl}} in Algorithm \ref{algo:FL}. The first initial cluster is equal to the whole population.\footnote{The presented algorithm assumes sampling the whole population each round. However, the modification taking into account dynamic online learning (where clients appear and disappear on an ongoing basis) is possible in several ways. This is explored in the last section of this article.} The first $T_{-1}$ value is set to $-inf$ (line 3) because we expect the temperature to rise at the beginning of the training and then descend. This descent will suggest the beginning of the convergence process - the first suitable moment for performing clustering. Since we are considering one-shot clustering, the flag $\mathcal{F}$ is initially set to false (line 4). Lines 5-13 describe the beginning of the standard \fl process, with computation of local stochastic gradient (line 9), performing local update (line 10) and computing local model delta (line 12), all done in parallel. After the gradients are aggregated, the next step is to check whether the clustering was already performed (line 14). If not, the matrix $\Gamma$ (line 12) capturing the cosine distance between the gradients is reconstructed, and the \textbf{Clustering Temperature} (eq. \eqref{eq: Temperature}) is calculated (line 17). If the newly registered temperature exceeds the one previously registered (line 18), we perform an actual clustering and switch the flag $\mathcal{F}$ to $True$. Otherwise, we simply return the initial clustering structure (line 23). The in-cluster models are reconstructed from gradients of clients attributed to a specific cluster (lines 26-29). Please note that for clarity, we have replaced the symbol of $\boldsymbol{\Theta}$ with a symbol $\theta$ in the description of the algorithm.

\begin{algorithm}
    \caption{One-Shot Clustered Federated Learning (\ocfl)}
    \label{algo:FL}
    \DontPrintSemicolon
    \SetKw{Input}{input}
    \Input{$\mathcal{A}(\cdot)$, Clustering Algorithm} \;
    \Input{$C = \{ P \}$, Set of Clusters}\;
    $T_{-1} \longleftarrow -inf$ \;
    $ \mathcal{F} \longleftarrow False$ \;
    \For{$ t = 0,...,T $}{        
        \For{client $ i \in P$ in parallel}{
            $\theta^{(t, 0)}_i \longleftarrow \theta^{t}$\;
            \For{$k = 0,...,K$}{
                $g_i( \theta_i^{(t, k)}) \longleftarrow \nabla \mathcal{L}(h(D_i, \theta_i^{(t, k)}), y)$ \;
                $\theta^{(t, k+1)}_i = $ ClientOpt $ (\theta_i^{(t, k)}, g_i(\theta_i^{(t, k)}), \tau) \;$
            }
            $\Delta_i^{(t)} \longleftarrow \theta_i^{(t, k+1)} - \theta_i^{(t, 0)}$\;
        }
        \If{$\mathcal{F} == False$}{
            $ \boldsymbol{\Gamma}^{(t)} \longleftarrow C_d(\Delta^{(t)}) $ \;
            $\lambda \longleftarrow \sqrt[\uproot{1} p]{n (n - 1) 2^p}$ \;
            $ T_{t} \longleftarrow \frac{||\boldsymbol{\Gamma}^{(t)}||_{p}}{\lambda} $ \;
            \eIf{
            $ T_{t} \geq T_{t-1} $ \;} { 
            $ C \longleftarrow \mathcal{A}(\boldsymbol{\Gamma}) $ \;
        $ \mathcal{F} \longleftarrow True $}
        {$C \longleftarrow C$}}
        \For{$C_j \in C$}{
            $ \Delta_j^{(t)} \longleftarrow \{ \Delta_i^{t} | i \in C_j \} $ \;
            $ g_{C_j} = \frac{1}{|C_j|}\sum_{k \in \Delta_j^{t}} \Delta_K^{t}$ \;
        }
        }
\end{algorithm}

\section{Experiments}
\label{sec:experiments}
In this section, we present a number of experiments evaluating the performance of \ocfl clustering algorithm (subsection \ref{experiment: clustering_correctness}), the impact of the clustering on the personalization (performance on the local test set) and generalization (performance on the hold-out test set) (subsection \ref{experiment: clustering_performance}) and the behaviour of the \textit{Clustering Temperature} Function (subsection \ref{experiment: temperature_function}). The data on which the tests were evaluated is reported at the beginning of this section (subsection \ref{experiment: datasets}), and the experimental setting is reported in subsection \ref{experiment: set-up}.

\subsection{Datasets} \label{experiment: datasets}
We experiment on three different image classification datasets with increased levels of difficulty: MNIST \cite{b15}, FMNIST \cite{b16}, and CIFAR10 \cite{b17}. We use the methodology described in subsection \ref{methodology: data_process} to simulate four different types of data splits: overlapping balanced, non-overlapping balanced, overlapping imbalanced and non-overlapping imbalanced for 15 and 30 clients.  The combination of three datasets, four different splits and two cardinalities of clients provide for a total of 24 different scenarios.

The \textbf{non-overlapping balanced splits} resemble the most basic case, where the cluster has access to separate classes, the agents are uniformly distributed across the clusters (with the same number of agents belonging to each cluster), and the available labels are uniformly distributed across the clients. The \textbf{overlapping balanced split} implies similar circumstances but with a partial overlap of classes, i.e., the clients partially share some of the classes. Imbalance is simulated across two different levels simultaneously. Firstly, clients are unevenly distributed across clusters. The first cluster holds $20 \% $ of the clients, while the second and third are approximately $47 \%$ and $33 \%$, respectively. Secondly, the distribution across labels is not uniform but is sampled according to weights vector $\sim Dir(\alpha)$, with parameter alpha set to 1. \textbf{The non-overlapping imbalanced split} assumes that the data is distributed according to $Dir(\alpha)$, but the set of classes is still disjoint across the clusters. On the contrary, overlapping an imbalanced split allows for a partial overlap. Referring to the methodology presented in subsection \ref{methodology: data_process}, the four different splits simulate four distinct cases that may be theoretically modelled by using different data-generating distributions.

\subsection{Experimental Set-up} \label{experiment: set-up}
Given that our proposal is clustering algorithm agnostic, we combine it with the density-based clustering methods such as MeanShift \cite{b19} and HDBSCAN \cite{b20} or parameter-free methods such as Affinity Propagation \cite{b21}. We employ four different baselines to test out the algorithm's general capabilities. The first one is \textbf{Baseline No-Clustering (BNC)}, which implies keeping all the clients in one cohort, as it is done in vanilla \fl. The next two baselines are derived from the literature, namely \textbf{Clustered Federated Learning: Model-Agnostic Distributed Multitask Optimization Under Privacy Constraints (SCL)} from \cite{b7} and \textbf{Federated learning with hierarchical clustering of local updates to improve training on non-IID data (BCL)} from \cite{b1}. The fourth baseline measures the performance of our algorithm in an ideal state, that is, a combination of \textbf{\ocfl with K-Means} where the number of data-generating distributions is known \textit{a priori}. Those four baselines are then compared against the combination of \ocfl with HDBSCAN (\textbf{OCFL-HDBSCAN}), Mean-shift (\textbf{OCFL-MeanShift}) and Affinity Propagation (\textbf{OCFL-Affinity}). The following section presents the experiments carried out to assess their efficiency in comparison to other algorithms.

To solve MNIST and FMNIST tasks locally, we employ a Deep Convolutional Neural Network of our own architecture inspired by tiny-VGG \cite{b28}. To solve CIFAR10, we employ ResNet34 \cite{b18} architecture. According to the experiments performed on the centralised version of the datasets, we were able to achieve a $99 \%$ and $94 \%$ test accuracy on MNIST and FMNIST tasks, respectively, and around $82 \%$ of accuracy on the CIFAR10 test set. To rule-out the possibility that excessive hyperparameters fine-tuning will influence the clustering results, we employed Stochastic Gradient Descent (SGD) and a shared local learning rate of $0.01$ for all local optimizers.

We employ a baseline \fedopt algorithm described in Section \ref{RelatedWork: FL} for the Federated Learning part. While the method can be easily adapted to other algorithms described in section \ref{RelatedWork: FL}, \fedopt (that is a generalization of \fedavg) can serve as a stable and easy-to-compare baseline. In our library, we have implemented algorithms of \cite{b1} and \cite{b7} from scratch, using pseudo-algorithms provided in their papers as well as publicly available code in the case of \cite{b7}.\footnote{\href{https://github.com/felisat/clustered-federated-learning}{Using the following GitHub Repository.}} Since the authors' implementation of \cite{b7} included in the GitHub repository contained an additional hyperparameter - a minimal number of rounds after which clustering can be performed - we have also transferred that hyperparameter into our implementation. We have performed a limited grid search for the algorithm of Sattler \etal \cite{b7} fixing the search intervals $e_1 \in [0.2, 0.6]$ and $e_2 \in [1.2, 2.0]$ for both parameters. If that failed, we turned it into a heuristic-based search. For the algorithm presented in \cite{b7}, we found a working combination for values of $e_1 = 0.35$, $e_2=0.35$ for MNIST and FMNIST tasks. For the algorithm reported in \cite{b1}, we experimented with a distance threshold in the $[0.05, 0.4]$ range, cosine similarity metric and linkage by averaging. We kept only the best-on-average results.

For each split of the dataset, we run 50 (MNIST, FMNIST) or 80 (CIFAR-10) global training rounds, where one round is preceded by three local epochs. For MNIST and FMNIST, the local batch size is set to 32. For CIFAR-10 local batch is set to 64. Since the HDBSCAN requires a minimal cluster size and we wanted to keep our clustering method hyper-parameter free, the minimal cluster size is always set to $20 \%$ of the sample size. The full archive of numerical values obtained during experiments, together with the code necessary to re-create the results and their visualization, was stored as a GitHub repository and is attached to this article. \footnote{\href{https://github.com/MKZuziak/OCFL_IEEE_BigData2024}{Link to the GitHub repository.}}

\subsection{Clustering Evaluation} \label{experiment: clustering_correctness}
We evaluate clustering by measuring the ability of the algorithm to attribute each client to its data-generating cluster correctly. Since the presented notion of \cfl is wholly based on the number of pre-defined data-clustering distributions (as described in subsection \ref{methodology: data_process}), we can test out the performance of an original bijective function $\mathcal{T}: C \longrightarrow \mathcal{P}$ attributing subset of clients to a certain cluster and the returned function $\hat{\mathcal{T}}$. Given the population $P = \{p_1, p_2, p_3, \cdots p_n \}$, we can define the original partition of $P$ into $n$ clusters, i.e. $C = \{C_1, C_2, \cdots, C_n \}$ and the automatically detected partition of $P$ into $n$ subsets, i.e. $\hat{C} = \{ \hat{C}_1, \hat{C}_2, \cdots, \hat{C}_m \}$. The closer the former resembles the primer, the better the formal clustering correctness of the presented solution. To evaluate the presented clustering methods, we employ \textbf{Rand Index (RAND)}, \textbf{Adjusted Mutual Information (AMI)} and \textbf{Completeness Score (COM)}.

\begin{figure*}
    \centering
cx    \includegraphics[width=1\linewidth]{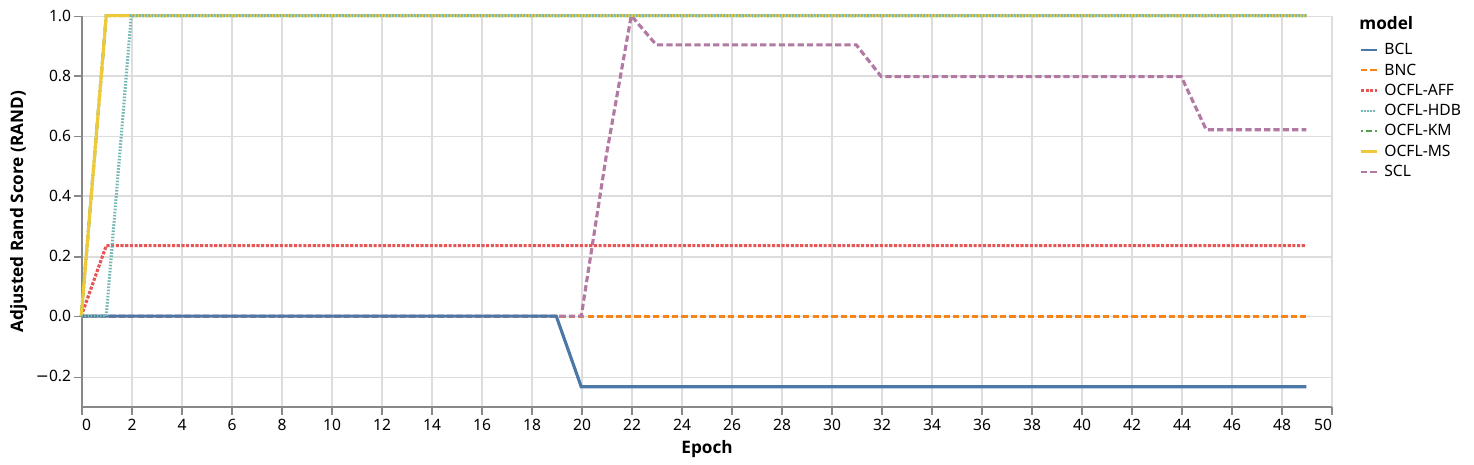}
    \caption{Exemplary figure presenting a continuous clustering assessment. Iterations are plotted on the x-axis, while the Adjusted Rand Score (RAND) is plotted on the y-axis. The simulation presents only a single run for the FMNIST task on a nonoverlapping balanced setting for 15 clients. The aggregated values for each possible combination of task, split and number of clients are presented in Table \ref{tab:experiment_EC: aggregated_clustering}}.
    \label{fig:experiment_EC: FMNIST_clustering_example}
\end{figure*}

The method of evaluation is as follows: for each of the experiments, all three of the scores are calculated between clustering $C$ and $\hat{C}$. Computing the score each round rewards clustering algorithms that are able to detect the correct structure at an early stage of the learning. On the other hand, it penalizes methods that either perform clustering at a too-early stage (resulting in detecting an improper structure) or grind the population structure into too many distinct clusters. The exemplary progression of clustering scores is displayed in Figure \ref{fig:experiment_EC: FMNIST_clustering_example}. Aggregated results for all three datasets across every possible split are plotted in Table \ref{tab:experiment_EC: aggregated_clustering}.

\begin{table*}

\centering
\caption{Clustering performance measured in terms of Rand Index (RAND), Adjusted Mutual Information Score (AMI) and Completeness Score (COM) for all datasets (with number of clients equal to 15 or 30), data splits, and across seven different clustering algorithms: Baseline (BNC), Sattler \etal (SCL) \cite{b7}, Briggs \etal (BCL) \cite{b1} with K-Means Algorithm (OCFL-KM), with Affinity Algorithm (OCFL-AFF), with MeanShift algorithm (OCFL-MS) and OCFL with HDBSCAN algorithm (OCFL-HDBS). The triple hyphen (---) is placed if the algorithm has failed to perform any clustering in the given scenario.}
\begin{tabular}{cccccccccccccc}
\tiny
&&\multicolumn{6}{c}{NonOverlapping} & \multicolumn{6}{c}{Overlapping} \\
&&\multicolumn{3}{c}{Balanced} & \multicolumn{3}{c}{Imbalanced} & \multicolumn{3}{c}{Balanced} & \multicolumn{3}{c}{Imbalanced} \\
&& RAND & AMI & COM & RAND & AMI & COM & RAND & AMI & COM & RAND & AMI & COM \\
\toprule
\multirow{6}{*}{MNIST 15} & SCL & 0.44 & 0.46 & 0.57 & 0.57 & 0.57 & 0.57 & 0.21 & 0.31 & 0.57 & 0.29 & 0.38 & 0.57 \\
& BCL& --- & --- & --- & --- & --- & --- & --- & --- & --- & --- & --- & --- \\
& OCFL-KM& \textbf{0.96} & \textbf{0.96} & \textbf{0.96} & \textbf{0.96} & \textbf{0.96} & \textbf{0.96} & \textbf{0.96} & \textbf{0.96} & \textbf{0.96} & \textbf{0.96} & \textbf{0.96} & \textbf{0.96} \\
& OCFL-AFF& 0.23 & 0.27 & 0.26 & 0.13 & 0.058 & 0.12 & 0.23 & 0.27 & 0.26 & 0.3 & 0.3 & 0.3 \\
& OCFL-MS& 0.71 & 0.67 & \textbf{0.96} & \textbf{0.96} & \textbf{0.96} & \textbf{0.96} & 0.87 & 0.87 & \textbf{0.96} & \textbf{0.96} & \textbf{0.96} & \textbf{0.96} \\
& OCFL-HDB& \textbf{0.96} & \textbf{0.96} & \textbf{0.96 }& \textbf{0.96} & \textbf{0.96} & \textbf{0.96} & \textbf{0.96} & \textbf{0.96} & \textbf{0.96} & \textbf{0.96} & \textbf{0.96} & \textbf{0.96} \\
\cmidrule(lr{0.150em}){3-5}\cmidrule(lr{0.150em}){6-8}\cmidrule(lr{0.150em}){9-11}\cmidrule(lr{0.150em}){12-14}
\multirow{6}{*}{MNIST 30} & SCL& 0.45 & 0.49 & 0.57 & 0.57 & 0.58 & 0.57 & 0.19 & 0.3 & 0.57 & 0.26 & 0.33 & 0.57 \\
& BCL& --- & --- & --- & -0.044 & -0.067 & 0.033 & --- & --- & --- & --- & --- & --- \\
& OCFL-KM& \textbf{0.96} & \textbf{0.96} & \textbf{0.96} & 0.96 & 0.96 & 0.96 & 0.9 & 0.9 & 0.9 & \textbf{0.94} & \textbf{0.94} & \textbf{0.94} \\
& OCFL-AFF& 0.38 & 0.44 & 0.37 & 0.66 & 0.63 & 0.65 & 0.36 & 0.42 & 0.35 & 0.5 & 0.49 & 0.42 \\
& OCFL-MS& 0.85 & 0.84 & 0.94 & 0.94 & 0.94 & 0.94 & 0.87 & 0.87 & 0.92 & \textbf{0.94} & \textbf{0.94} & \textbf{0.94} \\
& OCFL-HDB& 0.92 & 0.92 & 0.92 & \textbf{0.98} & \textbf{0.98} & \textbf{0.98} & \textbf{0.94} & \textbf{0.94} & \textbf{0.94} & \textbf{0.94} & \textbf{0.94} & \textbf{0.94} \\
\cmidrule(lr{0.150em}){3-5}\cmidrule(lr{0.150em}){6-8}\cmidrule(lr{0.150em}){9-11}\cmidrule(lr{0.150em}){12-14}
\multirow{6}{*}{FMNIST 15} & SCL& 0.46 & 0.48 & 0.57 & 0.55 & 0.53 & 0.57 & 0.19 & 0.28 & 0.57 & 0.17 & 0.27 & 0.57 \\
& BCL& -0.14 & -0.23 & 0 & --- & --- & --- & --- & --- & --- & --- & --- & --- \\
& OCFL-KM& \textbf{0.98} & \textbf{0.98} & \textbf{0.98} & \textbf{0.98} & \textbf{0.98} & \textbf{0.98} & \textbf{0.98 }& \textbf{0.98} & \textbf{0.98} & \textbf{0.98 }& \textbf{0.98} & \textbf{0.98} \\
& OCFL-AFF& 0.23 & 0.27 & 0.27 & 0.22 & 0.25 & 0.26 & 0.23 & 0.27 & 0.26 & 0.3 & 0.31 & 0.31 \\
& OCFL-MS& \textbf{0.98} & \textbf{0.98} & \textbf{0.98} & \textbf{0.98} & \textbf{0.98} & \textbf{0.98 }& \textbf{0.98} & \textbf{0.98} & \textbf{0.98} & \textbf{0.98} & \textbf{0.98} & \textbf{0.98} \\
& OCFL-HDB& 0.96 & 0.96 & 0.96 & \textbf{0.98} & \textbf{0.98} & \textbf{0.98} & 0.96 & 0.96 & 0.96 & \textbf{0.98} & \textbf{0.98} & \textbf{0.98} \\
\cmidrule(lr{0.150em}){3-5}\cmidrule(lr{0.150em}){6-8}\cmidrule(lr{0.150em}){9-11}\cmidrule(lr{0.150em}){12-14}
\multirow{6}{*}{FMNIST 30} & SCL& 0.57 & 0.57 & 0.57 & 0.57 & 0.58 & 0.57 & 0.42 & 0.43 & 0.57 & 0.53 & 0.52 & 0.57 \\
& BCL& --- & --- & --- & --- & --- & --- & -0.07 & -0.15 & 0 & -0.047 & -0.089 & 0.0036 \\
& OCFL-KM& 0.96 & 0.96 & 0.96 & \textbf{0.98} & \textbf{0.98} & \textbf{0.98} & \textbf{0.96 }& \textbf{0.96} & \textbf{0.96} & 0.96 & 0.96 & 0.96 \\
& OCFL-AFF& 0.39 & 0.45 & 0.37 & 0.66 & 0.64 & 0.67 & 0.37 & 0.43 & 0.36 & 0.51 & 0.5 & 0.43 \\
& OCFL-MS& 0.85 & 0.84 & \textbf{0.98} & 0.96 & 0.96 & 0.96 & 0.94 & 0.94 & 0.94 & \textbf{0.98} & \textbf{0.98} & \textbf{0.98} \\
& OCFL-HDB& \textbf{0.98 }& \textbf{0.98} & \textbf{0.98} & \textbf{0.98} & \textbf{0.98} & \textbf{0.98} & 0.94 & 0.94 & 0.94 & 0.96 & 0.96 & 0.96 \\
\cmidrule(lr{0.150em}){3-5}\cmidrule(lr{0.150em}){6-8}\cmidrule(lr{0.150em}){9-11}\cmidrule(lr{0.150em}){12-14}
\multirow{6}{*}{CIFAR 15} & SCL& --- & --- & --- & --- & --- & --- & --- & --- & --- & --- & --- & --- \\
& BCL& 0 & 1.7e-15 & 0.6 & 0 & 3.2e-16 & 0.6 & 0 & 1.7e-15 & 0.6 & 0 & 3.2e-16 & 0.6 \\
& OCFL-KM& 0.35 & 0.4 & 0.43 & 0.52 & 0.6 & 0.6 & \textbf{0.9} & \textbf{0.9} & \textbf{0.9} & 0.68 & 0.67 & 0.66 \\
& OCFL-AFF& --- & --- & --- & --- & --- & --- & 0.084 & 0.11 & 0.21 & 0.35 & 0.32 & 0.31 \\
& OCFL-MS& 0.23 & 0.26 & 0.57 & 0.16 & 0.18 & 0.58 & --- & --- & --- & 0.73 & 0.67 & 0.73 \\
& OCFL-HDB& \textbf{0.73} & \textbf{0.73} & \textbf{0.77} & \textbf{0.84 }& \textbf{0.84} & \textbf{0.84} & 0.59 & 0.59 & 0.69 & \textbf{0.77} & \textbf{0.77} & \textbf{0.74} \\
\cmidrule(lr{0.150em}){3-5}\cmidrule(lr{0.150em}){6-8}\cmidrule(lr{0.150em}){9-11}\cmidrule(lr{0.150em}){12-14}
\multirow{6}{*}{CIFAR 30} & SCL& --- & --- & --- & --- & --- & --- & --- & --- & --- & --- & --- & --- \\
& BCL& 0.025 & 0.033 & 0.55 & 0.021 & 0.029 & 0.55 & 0.0078 & 0.018 & 0.54 & 0.0047 & 0.015 & 0.54 \\
& OCFL-KM& 0.37 & 0.54 & 0.51 & 0.42 & 0.56 & 0.56 & \textbf{0.9 }& \textbf{0.9} & \textbf{0.9} & 0.59 & 0.55 & 0.58 \\
& OCFL-AFF& 0.35 & 0.41 & 0.34 & 0.46 & 0.45 & 0.39 & 0.23 & 0.22 & 0.2 & 0.32 & 0.3 & 0.26 \\
& OCFL-MS& 0.63 & 0.55 & \textbf{0.86} & 0.56 & 0.63 & 0.58 & --- & --- & --- & --- & --- & --- \\
& OCFL-HDB& \textbf{0.77} & \textbf{0.77} & 0.77 & \textbf{0.88 }& \textbf{0.88} & \textbf{0.88} & 0.64 & 0.64 & 0.69 & \textbf{0.83} & \textbf{0.83} & \textbf{0.81} \\
\bottomrule
\end{tabular}
\label{tab:experiment_EC: aggregated_clustering}

\end{table*}

\subsection{Local and Global Performance} \label{experiment: clustering_performance}
The local and global performance visualizes the trade-off between personalization (the ability of the model to perform well on local learning tasks) and generalization (the same ability but in relation to a broader set of examples). During this experiment, we define personalization capabilities as mean F1-score (PF1) averaged over all the client and training rounds. As noted before, the averaging values over all the iterations reward algorithms that can achieve better personalisation results at an early stage and penalize those that tend to run into sub-optimal solutions due to the incorrect clustering of clients. Moreover, it allows us to present results in one figure containing all the performed experiments. Generalization, on the other hand, is defined as a mean F1-score (GF1) achieved on the uniformly distributed test set of the orchestrator. To visualize the trade-off between personalization and generalization, we define a metric called learning gap (DIST) as follows:

\begin{equation}\label{eq: Dist}
    Dist = |PF1 - GF1|
\end{equation}

\begin{figure*}
    \centering
    \includegraphics[width=1\linewidth]{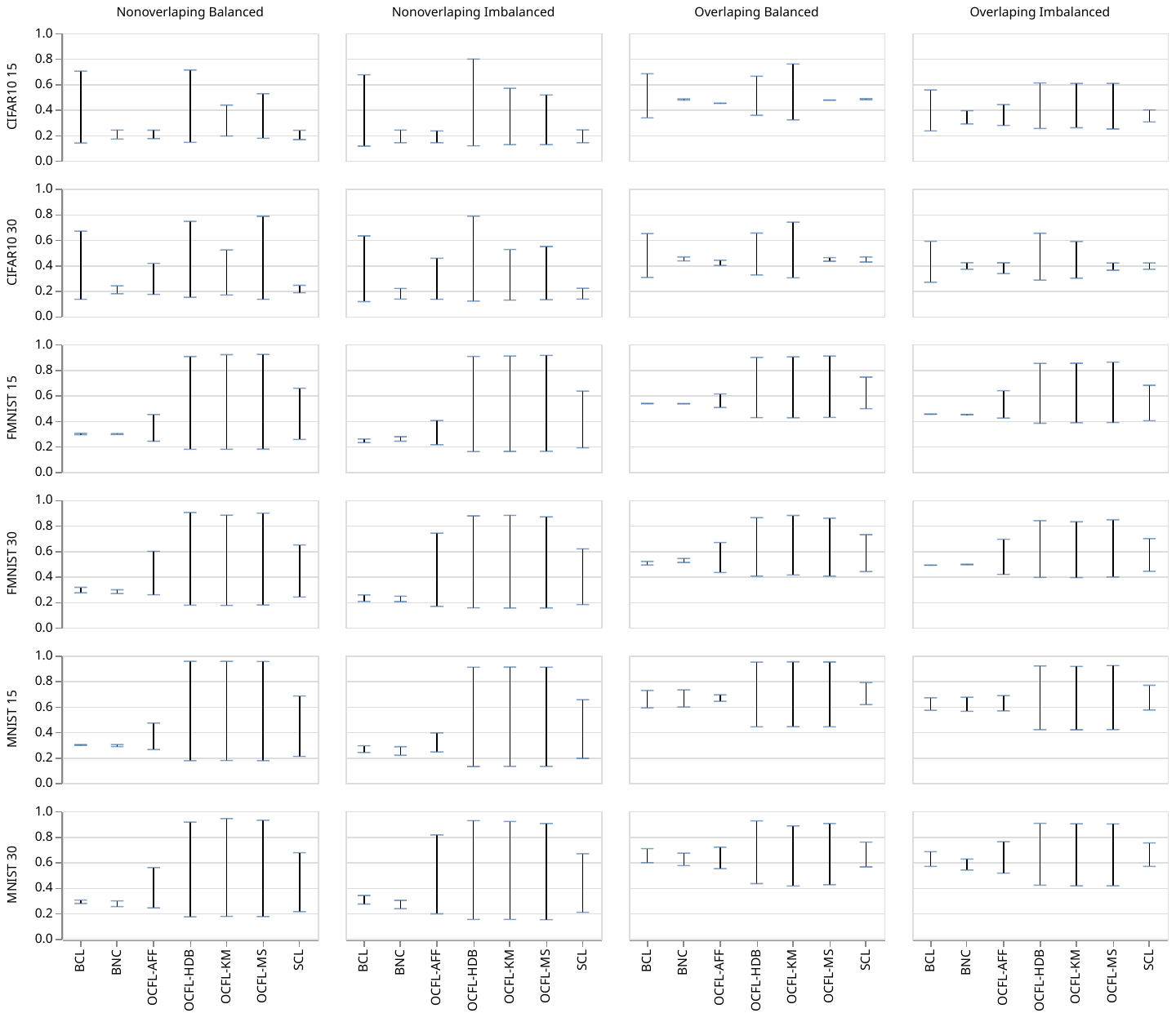}
    \caption{Aggregated performance of models in terms of F1-score. In the plot, the lower part of each segment indicates generalized F1-score (GF1), while the upper part indicates personalized F1-score (PF1). The length of the segment represents the learning gap as defined in Equation \ref{eq: Dist} }
    \label{fig:experiment: aggregated_performance}
\end{figure*}

We utilize the fact that, as described in the section \ref{experiment: datasets}, the orchestrator dataset consists of a held-out test dataset, which makes it a proper candidate for testing the generalization capabilities of the model. Ideally, the model should be characterized by a high F1-score achieved on a local dataset and a small learning gap. It would imply that the model has achieved high personalization capabilities while retaining a high degree of general knowledge. On the other hand, a low PF1 with a low DIST score would imply that the model has not been personalized. The results of the experiments are presented in Figure \ref{fig:experiment: aggregated_performance}.

\subsection{Behaviour of Temperature Function}\label{experiment: temperature_function}
\begin{figure*}
    \centering
    \includegraphics[width=1\linewidth]{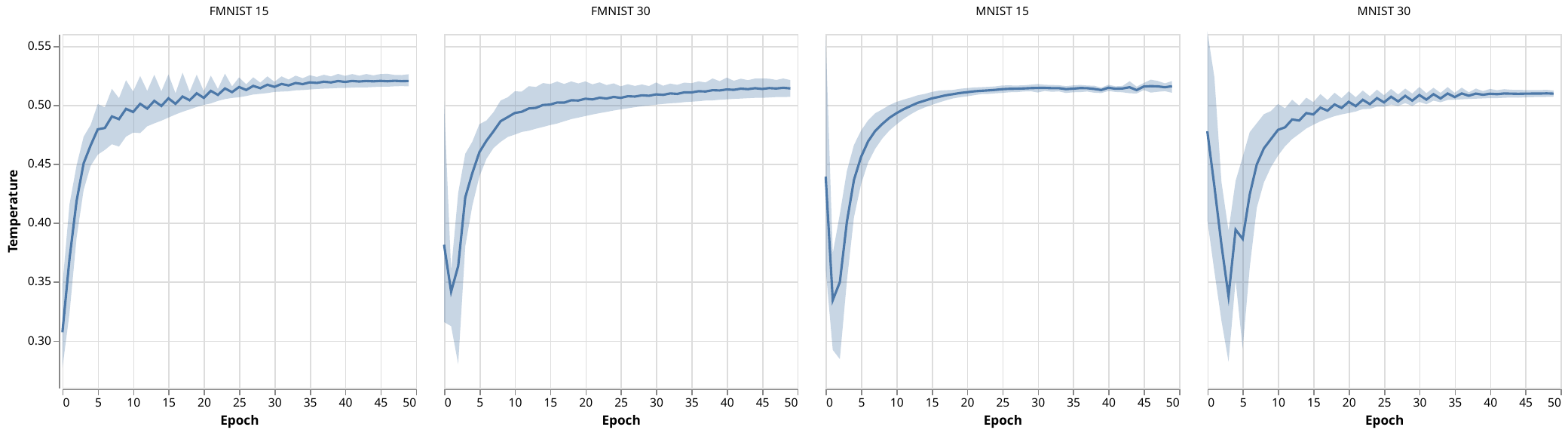}
    \caption{Behaviour of the temperature function as defined in the Equation \ref{eq: Temperature}. A solid blue line symbolizes the mean across all the scenarios (overlapping and nonoverlapping with balanced and imbalanced settings). Blue shadow represents confidence intervals established based on the variations between different scenarios.}
    \label{fig:experiment_AB: temperature_records}
\end{figure*}

In order to evaluate the behaviour of the temperature function (Equation \ref{eq: Temperature}) that is used as a method of detecting suitable clustering moment, we register the temperature of the baseline cluster each round for all the experiments performed on the MNIST and FMNIST tasks. Visualization of the experiments is reported in Figure \ref{fig:experiment_AB: temperature_records}. Each plot presents a stacked representation of each task in a given number of clients. A solid blue line symbolizes the mean across all the scenarios (overlapping and nonoverlapping in balanced and imbalanced settings). Blue shadow represents confidence intervals established based on the variations between different scenarios.

\subsection{Discussion} \label{experiment: discussion}
An overview of the experimental data can indicate the high performance of the \ocfl algorithm, especially when it is combined with density-based methods such as Meanshift or HDBSCAN. In terms of clustering performance (section \ref{experiment: clustering_correctness}), OCFL-HDB and OCFL-MS often outperform all other compared algorithms, with the RAND scores around 0.95 on all splits across the MNIST dataset, 0.96 across FMNIST dataset and 0.80 across the CIFAR10 dataset (Table \ref{tab:experiment_EC: aggregated_clustering}). Such a high RAND and Adjusted Mutual Information Score imply that both algorithms are able to almost perfectly distinguish between clusters at a very early stage, correctly separating cohorts of clients. Another key observation drawn from the data is the general difficulty of the distance-based algorithms, and it concerns state-of-the-art algorithms (such as \cite{b1} and \cite{b7}) and OCFL-AFF alike. Usage of those algorithms requires a proper hyper-tuning of hyperparameters, which is especially challenging in the federated environment. Although we followed the general guidelines presented in both articles and performed a reduced grid search reporting only the highest recorded values, the failure to cluster the clients into more than one or two clusters clearly indicates that those methods may often work well as a post-processing method. However, their use as a pre-processing (or in-training) clustering method may be problematic due to limited knowledge about the proper set of hyperparameters.\footnote{Indeed, there is a clear indication in both works of \cite{b7} and \cite{b3} (that we do not assess in this article) that the presented methods are a perfect solution for post-processing performed in the Federated setting.} There is also a clear correlation between SCL (\cite{b7}), BCL (\cite{b1}) and OCFL-AFF. In the cases where our limited grid search has returned a viable solution, those three algorithms tend to behave relatively well. However, when the grid search was initialized beyond the probable optimal region of search - the algorithms tended to run into multiple issues, with the total lack of clustering as the most severe one. 

The proper clustering should be reflected in the performance of personalized models, and this is clearly the case as presented in section \ref{experiment: clustering_performance}. The baseline model is characterized by a F1-score between the range of $0.24$ (CIFAR10 dataset as illustrated in \ref{fig:experiment: aggregated_performance}) and $0.6$ (for MNIST dataset). However, the local models delivered by the OCFL-MS and OCFL-HDB achieve F1-score between $0.96$ (MNIST) and $0.65$ (CIFAR10). What is important, their learning gap metric is often not much higher for local models trained by using OCFL-ML and OCFL-HDB, than it is for models trained using vanilla \fedopt. For example, local models trained for solving the CIFAR10 task in a cohort of 30 clients (overlapping imbalanced) have a learning gap that is higher only by $0.09$ from the baseline model while achieving an F1-score of $0.65$ (instead of 0.42). This means that clustering clients using that algorithm allows for more personalized local models that still retain similar generalization capabilities. Figure \ref{experiment: clustering_performance} can illustrate it quite well, as the generalized F1-score (lower part of each segment) is more or less comparable for all the presented methods, with the baseline serving as a most generalized case for obvious reasons. However, the personalized F1-score (higher part of each segment) is visibly better for the family of \ocfl methods, with particular emphasis on the high effectiveness of the OCFL-HDB.

The empirical behaviour of the temperature function reported in section \ref{experiment: temperature_function} indicates the properties that we mentioned in the earlier sections of the paper. In all the scenarios, we observed a drop in the temperature at the beginning of the training, then proceeded with a sharp rise as the gradients of the local models started to diverge due to incongruent local distributions. Around ten to twenty epochs after the start of the training, the temperature function stabilizes as the global model can not converge further. This behaviour is similar for all the presented datasets, except FMNIST in 15 clients split. However, in this case, there exists a minimal temperature peak between the first two global epochs. However, due to its minimal change, it is not clearly visible on the plot with the confidence intervals plotted.

It must also be highlighted that all four data splits generated for the experiments were challenging. Contrary to the label swaps or other similar methods often employed in studies of \cfl, our data splits assume more subtle differences in the data distribution as described in section \ref{experiment: datasets}. Such a subtlety may confound many clustering algorithms. Hence, a weaker performance of the OCFL-Kmeans or OCFL-Affinity - in line with the failures of algorithms presented in \cite{b1} and \cite{b7} to distinguish nodes and attribute them to more than one cluster. In this context, a high performance of density-based algorithms is even more valuable insight.

\section{Scope and limitations}
\label{sec:considerations}
\subsection{Privacy Issues}
\fl was proven to reduce the severity of unintended memorization - thus reducing the potential data leakage from aggregated models \cite{b29}. However, the baseline version of \fl is still vulnerable to privacy breaches, such as Membership Inference Attacks \cite{b30}, Reconstruction Attacks \cite{b31}, and cross-silo mitigation \cite{b41}. There are several strategies to mitigate those vulnerabilities, including Differential Privacy, which is well studied in the context of the \fl \cite{b32}. Due to space constraints, we have not focused on the privacy issue in this article, but our algorithm naturally extends to such a family of approaches as we do not execute any additional process on the client side nor aggregate any information other than gradients. Although the performance is expected to drop, DP can be easily applied to our method.

There is also a trade-off between the personalization degree and an attack vulnerability. It was proven that well-generalized models are less vulnerable when it comes to certain types of privacy attacks \cite{b33}. One of the advantages of proper clustering is the ability to create a correct clustering structure without creating a situation when some clients are singled out to a separate cluster. Such singled-out clusters can create a substantial privacy risk as they are training models that are overfitting a particular distribution. Due to space contains, we have focused on the clustering performance. However, subsequent studies should explore the relationship between clustering correctness and the success rate of privacy-related attacks.

\subsection{Dynamic Client Environment}
The \fl paradigm was designed to be dynamic, with clients joining and dropping out of the training at random intervals. In this paper, we have focused on a stable environment where all the clients are sampled each round. Since we use three different benchmark lines (baseline, the work of Sattler \etal \cite{b7} and the work of Briggs \etal \cite{b1}), we wanted to keep the environment stable, as to receive comparable and unbiased results. However, our algorithm naturally extends to a dynamic client environment, where One-Shot Clustering is performed on only a subset of clients. Depending on the chosen clustering algorithm, the subsequent client attribution will differ, but in most cases, it will rely on attributing each client to the most appropriate cluster by comparing the received gradients with the given clustering structure.  Depending on the ratio of the sample size to the population size, One-Shot Clustering can be turned into Few-Shots Clustering, where the clustering is performed a few times, each time updating the centroids or other appropriate clustering structure.

Similarly to \cite{b1, b7} that we use as a comparison, we started from the stable federated scenario to briefly extend it into a dynamic environment. Undeniably, clustering is more sensible in a stable environment when there is a clear advantage of establishing a correct division between the sample space. However, we wanted to extend this work with a notion of dynamic clustering to make it complete and sign new directions of the research.

\section{Conclusions}
\label{sec:conclusions}
In this paper, we have presented a family of \ocfl clustering agnostic algorithms that are suitable for performing the split of population at an early stage in an automatic manner. Moreover, by an extensive empirical evaluation of 36 different scenarios, we have evidenced the efficiency of the proposed solution, especially when combined with density-based clustering (i.e. OCFL-MS or OCFL-HDB). Proposed algorithms are able to correctly identify the correct structure of data-generating distributions and attribute clients to an appropriate cluster as early as in the first three rounds of the \fl process. This has a direct impact on the personalization of local models, as evidenced by the empirical results. There are a few research directions that should be taken in order to expand this work. One of them is directly connected to the privacy impact of the federated clustering. As explained in the final section, upholding a proper balance between personalization and generalization is crucial for retaining any privacy guarantees under the \fl scenario. However, due to the space constraint, we have to leave this point for future studies.

\bibliographystyle{IEEEtran.bst}
\bibliography{conference_101719}

\end{document}